\documentclass[journal]{IEEEtran}

\ifCLASSINFOpdf
\else
   \usepackage[dvips]{graphicx}
\fi
\usepackage{url}

\usepackage{graphicx}
\usepackage{wrapfig}
\usepackage{flushend}
\usepackage{booktabs}
\usepackage{multirow}
\usepackage{amsmath, amssymb}
\usepackage{cite}

\begin{document}

\title{MSDS: Deep Structural Similarity with Multiscale Representation}

\author{Danling Kang, Xue-Hua Chen, Bin Liu, Keke Zhang\textsuperscript{*}, Weiling Chen and Tiesong Zhao, \IEEEmembership{Senior Member, IEEE}
\thanks{Danling Kang, Weiling Chen and Tiesong Zhao are with the Fujian Key Lab for Intelligent Processing and Wireless Communication of Media Information, College of Physics and Information Engineering, Fuzhou University, Fujian 350108, China. E-mails: 2504780958@qq.com, weiling.chen@fzu.edu.cn, t.zhao@fzu.edu.cn.}
\thanks{Xue-Hua Chen is with the School of Computing and Information Science, Fuzhou Institute of Technology, Fuzhou 350506, China. E-mail: petterc@fit.edu.cn}
\thanks{Bin Liu is with School of Electronic Engineering, Fuzhou Inistitute of Technology, Fuzhou 350506, China. E-mail: liubin@fit.edu.cn. }
\thanks{Keke Zhang is with the School of Computer and Information Engineering (School of Artificial Intelligence), and also with the Henan Engineering Research Center of Intelligent Business \& Internet of Things Technology, Henan Normal University, Xinxiang 453007, China. *Corresponding author. E-mail: zhangkeke@htu.edu.cn. }
}
\markboth{Journal of \LaTeX\ Class Files, Vol. 14, No. 8, August 2015}
{Shell \MakeLowercase{\textit{et al.}}: Bare Demo of IEEEtran.cls for IEEE Journals}
\maketitle

\begin{abstract}
Deep-feature-based perceptual similarity models have demonstrated strong alignment with human visual perception in Image Quality Assessment (IQA). However, most existing approaches operate at a single spatial scale, implicitly assuming that structural similarity at a fixed resolution is sufficient. The role of spatial scale in deep-feature similarity modeling thus remains insufficiently understood. In this letter, we isolate spatial scale as an independent factor using a minimal multiscale extension of DeepSSIM, referred to as Deep Structural Similarity with Multiscale Representation (MSDS). The proposed framework decouples deep feature representation from cross-scale integration by computing DeepSSIM independently across pyramid levels and fusing the resulting scores with a lightweight set of learnable global weights. Experiments on multiple benchmark datasets demonstrate consistent and statistically significant improvements over the single-scale baseline, while introducing negligible additional complexity. The results empirically confirm spatial scale as a non-negligible factor in deep perceptual similarity, isolated here via a minimal testbed. 
\end{abstract}

\begin{IEEEkeywords}
Image quality assessment, full-reference IQA, deep structural similarity, multiscale representation, learnable weighting
\end{IEEEkeywords}

\IEEEpeerreviewmaketitle

\section{Introduction}

\IEEEPARstart{D}{eep-feature-based} perceptual similarity models have demonstrated strong alignment with human visual perception in image quality assessment (IQA). In particular, methods such as DeepSSIM \cite{zhang2025} compute structural similarity in pretrained feature spaces without additional training, achieving competitive performance with low complexity. 
\begin{figure}[t]
    \centering
    \includegraphics[width=0.9\linewidth]{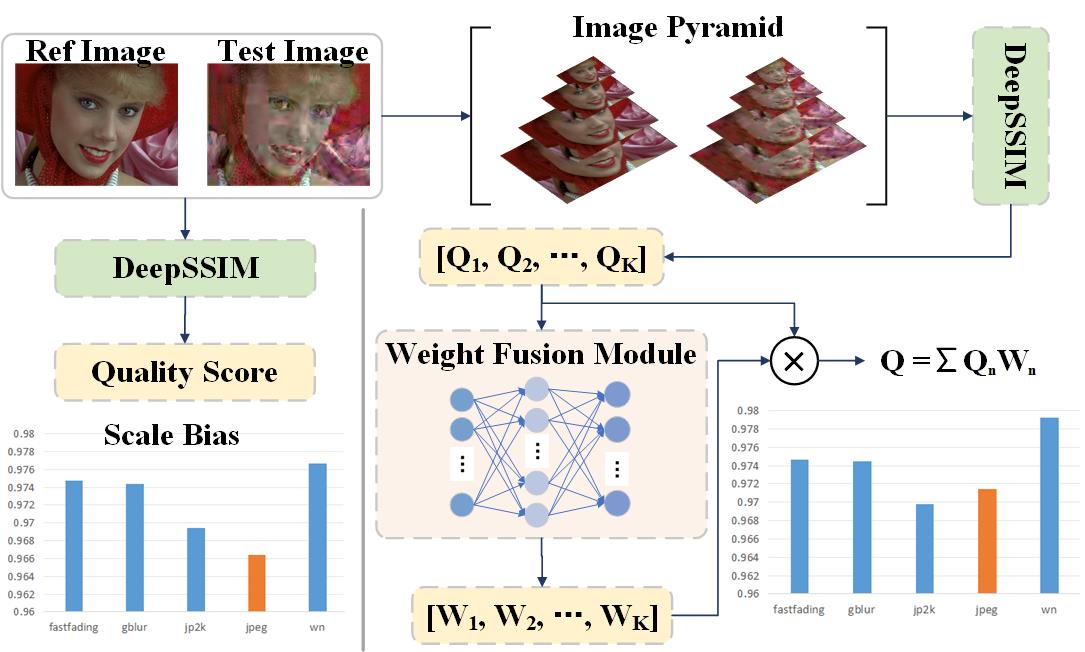}
    \caption{\textbf{Motivation of multiscale deep structural similarity. }Single-scale DeepSSIM assumes a fixed spatial scale is sufficient, introducing scale bias under frequency-diverse distortions. MSDS explicitly models deep structural similarity across multiple scales and integrates them for scale-aware perceptual quality assessment. }
    \label{fig:1}
\end{figure}

However, existing deep-feature-based IQA methods typically operate at a single spatial scale, implicitly assuming that structural similarity computed at a fixed resolution is sufficient. This assumption may introduce a scale bias, since different distortion types exhibit distinct frequency characteristics, as illustrated in Fig. \ref{fig:1}, and a single-scale representation cannot uniformly capture such cross-frequency effects, potentially limiting robustness. 

While multiscale modeling has been extensively studied in traditional IQA, such as MS-SSIM \cite{wang2003}, its role in deep-feature-based similarity remains insufficiently explored. Existing approaches often entangle feature representation and scale fusion, making it difficult to isolate the contribution of spatial scale.

In this letter, we present a Deep Structural Similarity with Multiscale Representation(MSDS), as a minimal and controlled multiscale extension of DeepSSIM to explicitly investigate spatial scale as an independent factor. The proposed framework computes DeepSSIM independently across pyramid levels and fuses the resulting scores via a lightweight set of learnable global weights, thereby decoupling perceptual representation from cross-scale integration. 

The contributions of this study can be summarized as follows:

\textit{1) }We isolate spatial scale as an independent factor via a minimal extension of DeepSSIM, decoupling feature representation and cross-scale integration to enable controlled analysis of scale effects in deep-feature-based IQA. 

\textit{2) }A small set of learnable global weights is introduced to aggregate scale-wise similarity scores, providing an interpretable and parameter-efficient mechanism without modifying pretrained feature representations. 

\textit{3) }Extensive experiments on multiple benchmark datasets demonstrate consistent and statistically significant improvements over the single-scale baseline (Wilcoxon signed-rank test, \(p < 0.05\)), supporting the relevance of spatial scale in deep perceptual similarity modeling. 

\begin{figure*}
    \centering
    \includegraphics[width=0.9 \textwidth]{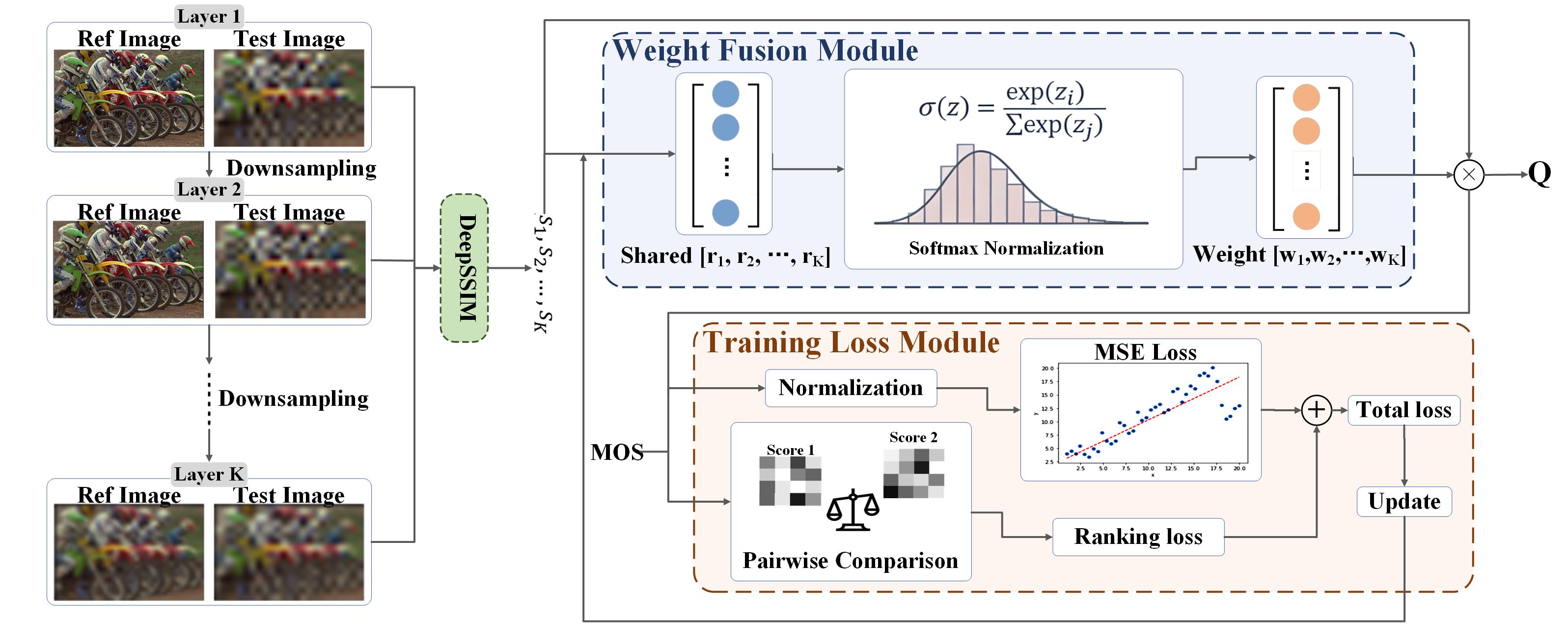}
    \caption{\textbf{Overall architecture of the proposed MSDS framework.} Given a reference image and a distorted image, Gaussian pyramids are first constructed to generate multiscale representations. At each scale, deep structural similarity is independently computed using the original DeepSSIM formulation. A small set of learnable global weights is then used to fuse the scale-wise similarity scores, yielding the final quality prediction. The loss function employs a standard MSE loss combined with a ranking loss.}
    \label{fig:overview}
\end{figure*}

\section{Related Work}
Early full-reference IQA methods rely on pixel-wise error estimation. Although computationally simple, these metrics correlate poorly with human subjective perception \cite{wang2006}. The Structural Similarity Index (SSIM) \cite{wang2004} proposed by Wang et al. assesses perceptual quality by modeling local image statistics, and its multi-scale extension, MS-SSIM \cite{wang2003}, demonstrates that aggregating information across spatial scales improves alignment with human perception. Subsequently, methods \cite{zhang2011, xue2014, sheikh2006a, ma2010, zhang2013, oszust2016} such as FSIM and GMSD further incorporate hand-crafted features and pre-defined fusion strategies. However, these approaches depend on manually designed features and fixed multi-scale fusion rules, which limits their generalization under complex distortion scenarios. 

With the development of deep learning, features extracted from pre-trained deep neural networks have become a mainstream choice for perceptual distance measurement. Methods \cite{johnson2016, ding2020, ding2024} such as LPIPS and DISTS employ deep features to characterize perceptual dissimilarity, while learning-based approaches like CONTRIQUE \cite{madhusudana2022} and $\rm IPS^2$ \cite{yan2025} further boost performance through large-scale training. In contrast, DeepSSIM directly computes structural similarity in a pre-trained feature space without requiring additional training, offering advantages in both efficiency and interpretability. However, these mainstream deep FR-IQA methods typically operate at a single spatial scale, implicitly assuming that a fixed receptive field is sufficient for perceptual modeling. 

Although multi-scale modeling has been extensively investigated in traditional IQA, its role in deep-feature-based similarity assessment still lacks systematic analysis. Several deep IQA works have attempted to incorporate multi-scale modeling: MILO \cite{cogalan2025} and BiRQA \cite{gushchin2026} adopt multi-scale architectures within FR-IQA frameworks, FsPN \cite{tang2025} and MS-SCANet \cite{mithila2025} introduce multi-scale features into NR-IQA, and LGDM \cite{saini2025} explores multi-scale perceptual representations within diffusion models. However, these approaches entangle feature representation, scale variation, and fusion strategy, making it difficult to isolate the specific contribution of the spatial scale variable. 

This work treats spatial scale as an independent modeling factor and proposes a minimal and controlled multi-scale extension of DeepSSIM. By decoupling feature representation from cross-scale fusion, the proposed framework enables a direct analysis of scale effects in deep perceptual similarity.

\section{Method}
\subsection{Overview}
We construct a minimal multiscale testbed based on DeepSSIM to empirically isolate the role of spatial scale in deep-feature-based IQA. The key idea is to decouple feature representation and scale integration by applying DeepSSIM independently across multiple spatial scales and aggregating the resulting similarity scores with a lightweight fusion mechanism. 

The overall pipeline is illustrated in Fig. \ref{fig:overview}. Given a reference image \(I_r\) and a distorted image \(I_d\), we first construct a multi-scale image representation. At each scale k, the similarity score \(s_k\) is computed using the original DeepSSIM. The final quality prediction is then obtained via cross-scale fusion.

\begin{table*}
\centering
\caption{Performance comparison of MSDS with state-of-the-art full-reference IQA methods across five benchmark datasets. }
\label{tab:results}
\begin{tabular}{c|cc|cc|cc|cc|cc}
\toprule
\multirow{2}{*}{Algorithms} & \multicolumn{2}{c}{LIVE} & \multicolumn{2}{c}{CSIQ} & \multicolumn{2}{c}{TID2013} & \multicolumn{2}{c}{KADID-10k} & \multicolumn{2}{c}{PIPAL} \\
\cmidrule(lr){2-3} \cmidrule(lr){4-5} \cmidrule(lr){6-7} \cmidrule(lr){8-9} \cmidrule(lr){10-11}
 & SRCC & PLCC & SRCC & PLCC & SRCC & PLCC & SRCC & PLCC & SRCC & PLCC \\
\midrule
SSIM \cite{wang2004} & 0.923 & 0.921 & 0.870 & 0.856 & 0.721 & 0.746 & 0.754 & 0.751 & 0.518 & 0.529 \\
FSIM \cite{zhang2011} & 0.963 & 0.960 & 0.917 & 0.903 & 0.802 & 0.859 & 0.832 & 0.829 & 0.605 & 0.638 \\
GMSD \cite{xue2014} & 0.960 & 0.960 & 0.950 & 0.947 & 0.804 & 0.859 & 0.843 & 0.843 & 0.609 & 0.656 \\
MS-SSIM \cite{wang2003} & 0.941 & 0.934 & 0.834 & 0.818 & 0.742 & 0.786 & 0.757 & 0.700 & 0.558 & 0.590 \\
\midrule
LPIPS \cite{zhang2018} & 0.932 & 0.936 & 0.884 & 0.906 & 0.673 & 0.756 & 0.721 & 0.713 & - & - \\
PieAPP \cite{prashnani2018} & 0.915 & 0.905 & 0.900 & 0.881 & 0.877 & 0.850 & 0.869 & 0.869 & - & - \\
DISTS \cite{ding2020} & 0.953 & 0.954 & 0.942 & 0.942 & 0.853 & 0.873 & - & - & - & - \\
CONTRIQUE-FR \cite{madhusudana2022} & \underline{0.966} & \textbf{0.966} & \underline{0.956} & \underline{0.964} & \textbf{0.909} & \textbf{0.915} & \textbf{0.946} & \textbf{0.947} & - & - \\
\midrule
DMM \cite{chen2025} & 0.947 & 0.913 & 0.946 & 0.917 & 0.876 & 0.879 & 0.918 & 0.913 & \textbf{0.729} & \textbf{0.752} \\
DeepSSIM \cite{zhang2025} & 0.964 & 0.959 & 0.950 & 0.948 & 0.838 & 0.860 & 0.903 & 0.902 & 0.699 & 0.734 \\
\midrule
MSDS & \textbf{0.967} & \textbf{0.966} & \textbf{0.964} & \textbf{0.967} & \underline{0.894} & \underline{0.911} & \underline{0.921} & \underline{0.921} & \underline{0.701} & \underline{0.736} \\
\bottomrule
\end{tabular}
\end{table*}

\subsection{Multiscale Similarity Computation}
The multi-scale representation is constructed via a standard Gaussian pyramid \cite{adelson1983} by progressively downsampling until a minimum resolution threshold is reached, which is set to 64 in this work. The number of scales \(K\) depends on the input image resolution, typically ranging from 3 to 4 in our experiments. 

At each scale, DeepSSIM is used to compute structural similarity. Specifically, a pretrained VGG-16 network is employed as a frozen feature extractor, taking features from its \(conv5\_1\) layer, and the similarity score \(s_k\) is computed based on the original DeepSSIM formulation. This design ensures that any performance variation stems solely from multi-scale modeling. 

\subsection{Cross-Scale Fusion}
To fuse the similarity scores from different scales, a lightweight set of learnable global weights is introduced. Let the maximum possible number of scales for a dataset be \(K\). A weight vector \(\mathbf{r}\in\mathbb{R}^K\) is normalized via Softmax to obtain \(w_k\). For images with fewer scales, only the first corresponding weights are used and re-normalized, while the similarity scores \(s_k\) for the missing scales are set to zero. The final quality score is:
\begin{equation}
\label{eq:final_score}
Q = \sum_{k=1}^K w_k \cdot s_k.
\end{equation}

This work adopts a lightweight linear fusion strategy, introducing only \(K\) learnable parameters. It preserves model interpretability while effectively mitigating overfitting under limited subjective data conditions. Moreover, this design does not alter the pretrained feature representations. 

\subsection{Training Strategy}
The fusion weights are optimized using a combination of normalized mean squared error and ranking loss to improve the consistency between predictions and subjective scores. Training employs the Adam optimizer, and the weights remain fixed during inference after convergence. Although the weights are learned on a per-dataset basis, the number of learnable parameters is only \(K\), making the training process stable and resistant to overfitting. Meanwhile, experiments show that the learned scale preferences exhibit a certain degree of generalization across different datasets.

\section{Experiments}
\subsection{Experimental Setup}
\textit{1) Datasets: }We evaluate the proposed method on five widely used full-reference IQA datasets, including LIVE \cite{sheikh2006b}, CSIQ \cite{chandler2010}, TID2013 \cite{ponomarenko2013}, KADID-10k \cite{lin2019}, and PIPAL \cite{jinjin2020}, covering both traditional and modern distortion types. 

\textit{2) Evaluation Metrics: }Performance is measured using Spearman rank-order correlation coefficient (SRCC) and Pearson linear correlation coefficient (PLCC) to assess prediction monotonicity and accuracy. 

\textit{3) Implementation Details: }Following standard practice, datasets are split by reference content into \(70\%\) training, \(10\%\) validation, and \(20\%\) testing. Experiments are repeated over 10 random splits, and median results are reported. The minimum resolution for multi-scale representation is set to 64 pixels, yielding \(K=3\sim4\) scales. Statistical significance is assessed using the two-sided Wilcoxon signed-rank test.

\subsection{Main Results and Statistical Significance}
Table \ref{tab:results} reports the performance comparison between the proposed method and representative full-reference IQA models. Compared with the single-scale DeepSSIM baseline, the proposed method achieves improvements across all five datasets. 

To verify the reliability of these gains, the Wilcoxon signed-rank test is conducted on the performance differences between MSDS and DeepSSIM, with results shown in Table \ref{tab:significance}. The improvements are statistically significant (\(p < 0.05\)) on all datasets.

\begin{table}[h]
\centering
\caption{Wilcoxon signed-rank test p-values for MSDS vs. DeepSSIM. (10 random splits)}
\label{tab:significance}
\begin{tabular}{ccc}
\toprule
Dataset & SRCC p-value & PLCC p-value \\
\midrule
LIVE & 0.02148 & 0.02148 \\
CSIQ & 0.02148 & 0.02148 \\
TID2013 & 0.00195 & 0.00195 \\
KADID10k & 0.00195 & 0.00195 \\
PIPAL & 0.0098 & 0.00195 \\
\bottomrule
\end{tabular}
\end{table}

Compared with learning-based models such as DISTS and CONTRIQUE-FR, the proposed method achieves competitive performance while introducing negligible additional parameters. This suggests that the observed improvements primarily stem from explicit multiscale modeling rather than increased model complexity.

\begin{table*}
\centering
\caption{Ablation study on multiscale modeling and cross-scale fusion strategies.}
\label{tab:ablation}
\begin{tabular}{c|cc|cc|cc|cc|cc}
\toprule
\multirow{2}{*}{Setting} & \multicolumn{2}{c}{LIVE} & \multicolumn{2}{c}{CSIQ} & \multicolumn{2}{c}{TID2013} & \multicolumn{2}{c}{KADID-10k} & \multicolumn{2}{c}{PIPAL} \\
\cmidrule(lr){2-3} \cmidrule(lr){4-5} \cmidrule(lr){6-7} \cmidrule(lr){8-9} \cmidrule(lr){10-11}
 & SRCC & PLCC & SRCC & PLCC & SRCC & PLCC & SRCC & PLCC & SRCC & PLCC \\
\midrule
Single-scale & 0.964 & 0.959 & 0.950 & 0.948 & 0.838 & 0.860 & 0.903 & 0.902 & 0.699 & 0.734 \\
Equal Weight & 0.964 & 0.961 & 0.925 & 0.918 & 0.871 & 0.885 & 0.882 & 0.881 & 0.685 & 0.735 \\
Fixed Weight & 0.950 & 0.947 & 0.886 & 0.877 & 0.843 & 0.858 & 0.814 & 0.811 & 0.622 & 0.691 \\
Minimum resolution=32 & 0.966 & 0.964 & 0.964 & 0.965 & 0.890 & 0.905 & 0.917 & 0.917 & 0.692 & 0.731 \\
Minimum resolution=128 & 0.966 & 0.965 & 0.963 & 0.965 & 0.894 & 0.905 & 0.921 & 0.921 & 0.695 & 0.733 \\
MSDS & 0.967 & 0.966 & 0.964 & 0.966 & 0.894 & 0.911 & 0.921 & 0.921 & 0.701 & 0.736 \\
\bottomrule
\end{tabular}
\end{table*}

\begin{table*}
\centering
\caption{Cross-database and cross-distortion generalization performance.}
\label{tab:cross}
\begin{tabular}{c|cc|cc|cc|cc}
\toprule
\multirow{2}{*}{Setting} & \multicolumn{2}{c}{LIVE} & \multicolumn{2}{c}{CSIQ} & \multicolumn{2}{c}{TID2013} & \multicolumn{2}{c}{PIPAL} \\
\cmidrule(lr){2-3} \cmidrule(lr){4-5} \cmidrule(lr){6-7} \cmidrule(lr){8-9}
 & SRCC & PLCC & SRCC & PLCC & SRCC & PLCC & SRCC & PLCC \\
\midrule
LPIPS \cite{zhang2018} & 0.940 & 0.939 & 0.881 & 0.899 & 0.394 & 0.471 & 0.613 & 0.654 \\
PieAPP \cite{prashnani2018} & 0.919 & 0.908 & 0.892 & 0.877 & 0.876 & 0.859 & 0.607 & 0.597 \\
DeepIQA \cite{bosse2018} & 0.947 & 0.940 & 0.909 & 0.901 & 0.831 & 0.834 & - & - \\
CONTRIQUE-FR \cite{madhusudana2022} & 0.898 & 0.889 & 0.843 & 0.855 & 0.630 & 0.665 & 0.470 & 0.465 \\
DISTS \cite{ding2020} & 0.954 & 0.954 & 0.929 & 0.928 & 0.830 & 0.855 & 0.655 & 0.687 \\
SSHMPQA \cite{xian2024} & \underline{0.963} & \underline{0.959} & \underline{0.945} & \underline{0.945} & \textbf{0.879} & \textbf{0.897} & \underline{0.692} & \underline{0.709} \\
MSDS & \textbf{0.967} & \textbf{0.964} & \textbf{0.955} & \textbf{0.953} & 0.864 & \underline{0.884} & \textbf{0.701} & \textbf{0.744} \\
\bottomrule
\end{tabular}
\end{table*}

\subsection{Distortion-Type Analysis}

\begin{table}[h]
\centering
\caption{Distortion-type analysis on the LIVE dataset. }
\label{tab:type}
\begin{tabular}{c|cc|cc}
\toprule
\multirow{2}{*}{Distortion type} & \multicolumn{2}{c}{DeepSSIM} & \multicolumn{2}{c}{MSDS} \\
\cmidrule(lr){2-3} \cmidrule(lr){4-5}
 & SRCC & PLCC & SRCC & PLCC \\
\midrule
JPEG & 0.960 & 0.973 & 0.967 & 0.976 \\
JPEG2000 & 0.964 & 0.975 & 0.964 & 0.976 \\
White Noise & 0.972 & 0.982 & 0.974 & 0.985 \\
Blur & 0.970 & 0.979 & 0.970 & 0.979 \\
Fastfading & 0.967 & 0.983 & 0.966 & 0.983 \\
\bottomrule
\end{tabular}
\end{table}

To investigate the effect of multiscale modeling, distortion-type analysis is conducted on the LIVE dataset, as summarized in Table \ref{tab:type}. 

Analysis shows that MSDS achieves modest but consistent improvements over DeepSSIM on several high-frequency-related distortions, such as JPEG compression and additive white noise, while performing nearly identically to DeepSSIM on low-frequency-dominated distortions such as blur. 

\begin{figure}[h]
    \centering
    \includegraphics[width=0.9\linewidth]{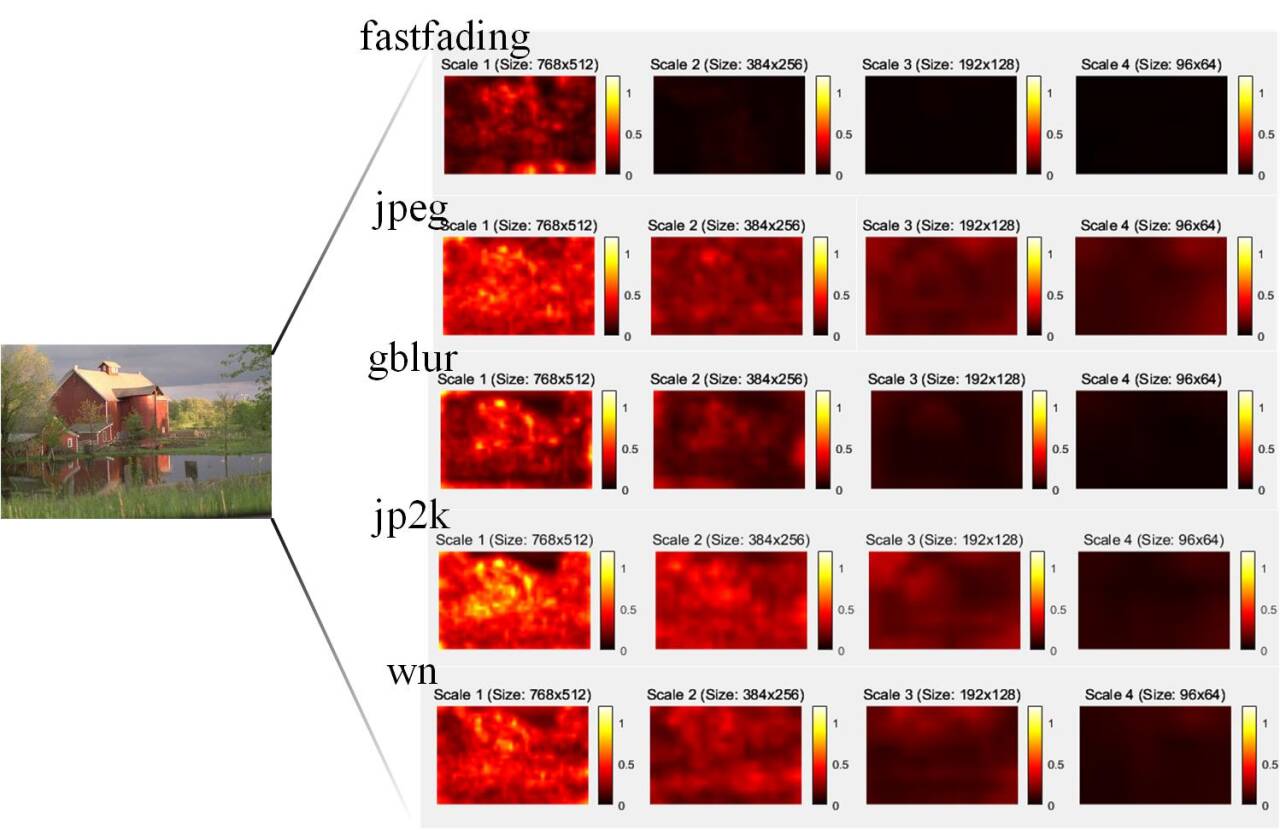}
    \caption{\textbf{Multiscale feature response maps for five distortion types (levels 1–4).} Brighter colors indicate larger structural discrepancies. The color range is fixed to [0, 1.2] across all maps. }
    \label{fig:feature}
\end{figure}

Fig. \ref{fig:feature} further visualizes the per-scale feature responses for five representative distortions. 

\subsection{Ablation Study}
An ablation study is conducted to analyze the effects of multi-scale modeling and fusion strategies. Results (Table \ref{tab:ablation}) show that introducing multi-scale representations consistently improves performance over the single-scale baseline.

Further comparison of different fusion strategies reveals that equal-weight fusion yields limited and unstable gains, while fixed weights adopted from MS-SSIM perform worse on most datasets. In contrast, the proposed learnable fusion strategy achieves the best performance, indicating the importance of adaptive scale fusion in deep-feature IQA.

\subsection{Generalization}
To evaluate generalization capability, the fusion weights learned on KADID-10k are directly applied to other datasets without any fine-tuning. Results (Table \ref{tab:cross}) demonstrate that the proposed method maintains stable performance across different datasets and distortion types.

Although the magnitude of improvement varies, the results indicate that the learned scale preferences exhibit a certain degree of generalization under distribution shift. Cross-dataset MSDS also improves upon the zero-shot DeepSSIM baseline. 

\section{Conclusion}
In this letter, we empirically isolated spatial scale as an independent factor in deep-feature-based image quality assessment using a minimal multiscale testbed. By decoupling feature representation and scale integration, the proposed framework enables a controlled analysis of scale effects. 

Experimental results demonstrate consistent and statistically significant improvements over the single-scale baseline with negligible additional complexity. These findings suggest that spatial scale is a relevant and non-negligible factor in deep perceptual similarity modeling, and can be effectively incorporated through a simple and interpretable design.


\end{document}